*Article*

# Recognition of Urbanized Areas in UAV-Derived Very-High-Resolution Visible-Light Imagery

**Edyta Puniach [1,*], Wojciech Gruszczyński [1], Paweł Ćwiąkała [1], Katarzyna Strząbała [1] and Elżbieta Pastucha [2]**

[1] AGH University of Krakow, Faculty of Geo-Data Science, Geodesy, and Environmental Engineering, Mickiewicza 30, 30-059 Krakow, Poland; wgrusz@agh.edu.pl (W.G.); pawelcwi@agh.edu.pl (P.Ć.); strzabal@agh.edu.pl (K.S.)
[2] The Mærsk Mc-Kinney Møller Institute, University of Southern Denmark, Campusvej 55, DK-5230 Odense, Denmark; elpa@mmmi.sdu.dk
* Correspondence: epuniach@agh.edu.pl

**Abstract:** This study compared classifiers that differentiate between urbanized and non-urbanized areas based on unmanned aerial vehicle (UAV)-acquired RGB imagery. The tested solutions included numerous vegetation indices (VIs) thresholding and neural networks (NNs). The analysis was conducted for two study areas for which surveys were carried out using different UAVs and cameras. The ground sampling distances for the study areas were 10 mm and 15 mm, respectively. Reference classification was performed manually, obtaining approximately 24 million classified pixels for the first area and approximately 3.8 million for the second. This research study included an analysis of the impact of the season on the threshold values for the tested VIs and the impact of image patch size provided as inputs for the NNs on classification accuracy. The results of the conducted research study indicate a higher classification accuracy using NNs (about 96%) compared with the best of the tested VIs, i.e., Excess Blue (about 87%). Due to the highly imbalanced nature of the used datasets (non-urbanized areas constitute approximately 87% of the total datasets), the Matthews correlation coefficient was also used to assess the correctness of the classification. The analysis based on statistical measures was supplemented with a qualitative assessment of the classification results, which allowed the identification of the most important sources of differences in classification between VIs thresholding and NNs.

## 1. Introduction

Image classification is a process that involves categorizing imagery pixels into meaningful classes or themes to help with interpreting and analyzing acquired imagery. One of the primary applications of image classification in remote sensing is the Land Use Land Cover (LULC) classification, which aims to map and monitor the Earth's surface. LULC classification can be employed both for large-scale data with a broad class range [1–3] as well as issue-specific tailored class solutions, like impervious surface classification [4–6].

Currently, most of the state-of-the-art methods rely on multi- and hyperspectral data, which are widely available thanks to satellite acquisitions [1–3]. These data facilitate LULC mapping on regional and global scales, offering a spatial resolution of several meters. Such resolution suffices for numerous tasks conducted at these scales, including urban planning, biodiversity research, or forestry management [3].

However, the spatial resolution of satellite-borne data can sometimes be too coarse to capture the spatial heterogeneity essential in many local-scale tasks. Moreover, the satellite revisit cycles, compounded by the influence of cloud cover, diminish the frequency of data acquisitions suitable for analysis. Conversely, the use of manned aerial vehicles, capable of capturing data at finer spatial resolutions (sub-meter), is often prohibitively expensive, especially for small areas. In such scenarios, unmanned aerial vehicles (UAVs) emerge as practical solutions for applications requiring high spatial resolutions, user-defined temporal resolution, low cost, and easy operability.

Most LULC classification methods using UAV-acquired data rely on multispectral sensors [7–9]. These sensors have lower geometric resolution and higher prices than their RGB counterparts. On the other hand, the accuracy of LULC classification achieved based on multispectral imagery is generally significantly higher than that using only RGB data [10]. Nevertheless, the paramount importance for many tasks lies in the high spatial resolution of the data provided by RGB cameras rather than their spectral resolution. An example of such a task is determining relatively small surface displacements, such as those induced by underground mining [11,12]. In such tasks, LULC classification may enhance result accuracy, but it is not deemed indispensable and is of lesser importance than the spatial resolution of acquired data. Consequently, RGB-based LULC classification often becomes the only feasible solution, as high-resolution multispectral data are unavailable.

LULC classification based on optical imaging can be performed through visual interpretation by an experienced interpreter. This approach's advantage is its high accuracy, generally surpassing that of computer-based classification methods. However, its main drawback is its time-consuming nature.

Another approach to LULC classification involves utilizing vegetation indices (VIs) and thresholding techniques [9]. VIs produce a single value per pixel by combining the responses from several spectral bands to enhance the presence of vegetation features in images. While multispectral VIs are commonly employed, the Normalized Difference VI (NDVI) is the most widely used [13,14]. Nonetheless, RGB-based VIs also demonstrate efficacy in LULC classification. For example, the Excess Green Minus Excess Blue (ExGB) VI was effectively utilized to classify RGB images into crop and non-crop classes for a crop harvesting video system [15].

Currently, machine learning (ML) and deep learning (DL) methods are widely used for LULC classification. ML methods, such as support vector machine (SVM) and random forest (RF), are characterized by high repeatability and time efficiency [9,16–18]. However, the classification accuracy often declines significantly when the data or study area changes [16]. On the other hand, DL methods have demonstrated exceptional performance in LULC classification [19], particularly when sufficient training samples are available [20]. To enhance classification accuracy, ancillary data—such as texture features, spatial and geometric features, and VIs—are sometimes integrated alongside the reflection values for the different spectral bands [21–25].

Only a few papers describe the usability of RGB images acquired with UAV for LULC classification or impervious surface classification. Thus, it remains an open research problem [10,26]. The solutions described so far differ widely. In one study [27], a convolutional neural network (CNN) was utilized to perform LULC classifications of UAV RGB images into seven classes. The classification was aided by a digital surface model (DSM) and achieved a high accuracy of up to 98%. Another study evaluated the performance of four CNN algorithms classifying karst wetland vegetation [24]. It demonstrated that using additional texture features and a DSM derived from UAV images improved the classification accuracies compared to using only RGB data. Semantic segmentation networks fed by UAV RGB images were also tested as LULC classifiers [19]. The study results showed that, compared with the ML method (AdaBoost), the three tested DL methods (U-Net, UNet++, and DE-UNet) achieve higher classification accuracy reaching up to 95%.

U-Net was also employed for urban land cover mapping using UAV-derived data [28]. In this study, six land cover classes were successfully classified with an accuracy of 91% using only RGB images. When RGB data was combined with Light Detection and Ranging (LiDAR) data, the classification accuracy improved, reaching 98%. The maximum likelihood and SVM methods were also applied for comparison, yielding similar results. However, U-Net outperformed both ML methods across all tested scenarios.

SVM was also applied in the classification of impervious surfaces into 16 classes, achieving an accuracy of 81% [29]. The authors propose expanding the number of classes from six in the base scheme in order to maximize the information obtained from UAV

RGB imagery and better address tasks related to urban space management. In another study on impervious surface classification [4], an adaptive boosting (AdaBoost) classifier was employed, utilizing randomized quasi-exhaustive multiscale features that captured both spectral and height information. The results showed that imperviousness maps generated from UAV RGB images achieved accuracy levels comparable to those derived from manned aerial RGB imagery. Specifically, AdaBoost achieved accuracies ranging from 93.7% to 96.2% for UAV data and from 95.6% to 97.4% for manned aerial data, while the maximum likelihood method, a well-known method, produced results that were up to 20% less accurate. Additionally, the study found that the number of target classes significantly influenced the performance of the ML method, with classifications involving three target classes being up to 9% less accurate than those with only two target classes.

UAV-derived RGB images combined with a DSM were also employed to create detailed maps of heterogeneous urban areas [23]. The proposed approach utilized geographic object-based image analysis (GEOBIA) alongside supervised multiscale ML methods. The study tested various features, ML methods, and scales during the classification of 12 classes, ultimately achieving accuracies as high as 94%.

In another study [30], ML methods (SVM, RF, and k-means) were assessed for classifying five distinct structures within UAV RGB images. The results demonstrated that the RF classifier outperformed the other methods achieving an accuracy of 92%. A similar conclusion was drawn in another study [22], where the RF classifier achieved the highest accuracy (~89%) among all tested methods. During the classification, 90 features were utilized, among which VIs proved to be the most relevant. The authors additionally demonstrated that the classification accuracy depended on the spatial resolution of RGB images and was lower for images originally acquired at the defined spatial resolution when compared to those resampled.

An SVM with a linear kernel was also used to perform LULC classification into three classes [10]. The authors compared the system's performance with RGB and multispectral data, with the latter achieving better results. In one study [26], a combination of RGB imagery and three different RGB-based VIs, the Triangular Greenness Index (TGI) [31], the Green Leaf Index (GLI) [32], and the Red Green Blue Vegetation Index (RGBVI) [33], were classified using four different classifiers, RF, XGBoost, LightGBM, and CatBoost. Datasets that included VIs showed significantly better results, reaching an accuracy of up to 92%. Similar conclusions were attained in another study testing multiple combinations of datasets and classifiers [34]. These datasets included combinations of RGB images, ten different VIs, and a height model. Among the tested classifiers were support vector regression, extreme learning machine, and RF.

In most of the research conducted so far [10,19,22,24,26,27,30,34], training and test datasets were the subsets of data acquired during the same UAV flights over the same study area. This implies that the estimated classification accuracy for the test datasets is strictly relevant to the particular study area. Furthermore, it does not provide insights into the generalizability of the developed LULC classification methods or how the classification accuracy might vary when altering data (i.e., utilizing data from a different UAV flight, with different spatial resolution, or during different seasons) or study areas. One study [28] briefly demonstrated this issue by analyzing data acquired from two study areas using different devices at different times. However, the study areas were quite small, and the results obtained require validation on larger datasets.

Moreover, discrepancies arise in the process of creating the reference classification required to train and test LULC classification methods developed so far. Two predominant approaches are commonly employed. The first involves constructing the reference dataset comprising tens to hundreds of randomly selected points or regions of interest [10,27,30]. In this case, ground truth information is acquired through visual interpretation of UAV-derived data, Google Earth images, or field investigation. Alternatively, another approach is based on surface reference classification, typically conducted through vectorization or manual semantic segmentation of RGB data [19,22,24]. However, this approach is generalized, and the spatial resolution of reference data is lower and does not align with that of the original RGB images.

The presented research studies are related to improving the system aimed at determining displacements based on UAV photogrammetry [11,12]. The valid system operation requires classification of a monitored region into two classes—urbanized and non-urbanized areas. Urbanized areas contain objects that do not change their appearance/texture over a long period and are fixed to the ground, i.e., changes in their positions can be assumed to be equal to ground surface displacement. For those areas, both horizontal and vertical displacement can be calculated [35]. Non-urbanized areas contain all other objects—mostly bare soil and vegetation. Only vertical displacement can be calculated here, usually with lower accuracy [36]. Thus, correct classification of a monitored region minimizes the number of wrongly determined horizontal displacements and accelerates data processing.

As a result, this research study aimed to find the most suitable pixel-wise classification method for this particular problem, that utilizes existing data (UAV RGB images only) and is simple to apply. The selection or development of such a method based on high-resolution images requires reference classification, which is both labor-intensive and time-consuming. Consequently, in practical applications, the calibration or training dataset is relatively small, which means that solutions with a low level of complexity (and degrees of freedom) seem more suited to this task. Therefore, the specific contributions of this study are as follows:

1. The utilized datasets comprise only high-resolution RGB images acquired from two distinct study areas. These images were acquired using two UAV platforms across multiple measurement series conducted at various stages of vegetation cover (different seasons), ensuring comprehensive validation and robustness of the tested classifiers;
2. The reference classification used to calibrate/train and test classification methods was meticulously manually performed at the full spatial resolution of RGB data, which was 10 mm and 15 mm, respectively;
3. The study undertook a comparative and evaluative analysis of 16 VIs, determined based on high-resolution UAV-derived RGB images, in conjunction with a simple thresholding approach, within the purview of the classification task at hand;
4. The influence of season and dataset on the optimal VIs' thresholds and classification accuracy was analyzed;
5. The classification accuracy using simple neural networks (NNs), including linear networks, multi-layer perceptron (MLP), and CNNs, trained especially for this purpose, was also assessed and compared with the classification results obtained based on VIs thresholding;
6. As input data for NNs consisted of a pixel and its surroundings (an image patch), the study also provided some feedback about neighborhoods' influence on the classification results.

This study is structured as follows. Section 2 presents the VIs used in the research project and the NNs with important parameters regarding their training, as well as the datasets used in this study. Section 3 presents the results of the classifications performed. Section 4 discusses the results in the context of literature research studies, encountered problems, and alternative solutions to those described in previous sections. Section 5 contains synthetic conclusions based on the conducted research.

## 2. Materials and Methods

### *2.1. Datasets*

The research datasets were acquired independently using two different UAVs and cameras at different spatial resolutions and in different weather conditions. This ensures the datasets' independence and, hence, the robustness of the assessment of the compared classification solutions.

#### 2.1.1. Jerzmanowice Dataset

Data used to optimize classification method parameters were acquired using a UAV in the central part of Jerzmanowice village (50°21′N, 19°76′E) in the Lesser Poland Voivodeship, Poland. The total area covered by UAV photogrammetry missions was 1.5 km$^2$ (Figure 1a). Single-family houses and farm buildings are concentrated along the streets. The remainder of the area is covered by agricultural land (mostly arable land and meadows).

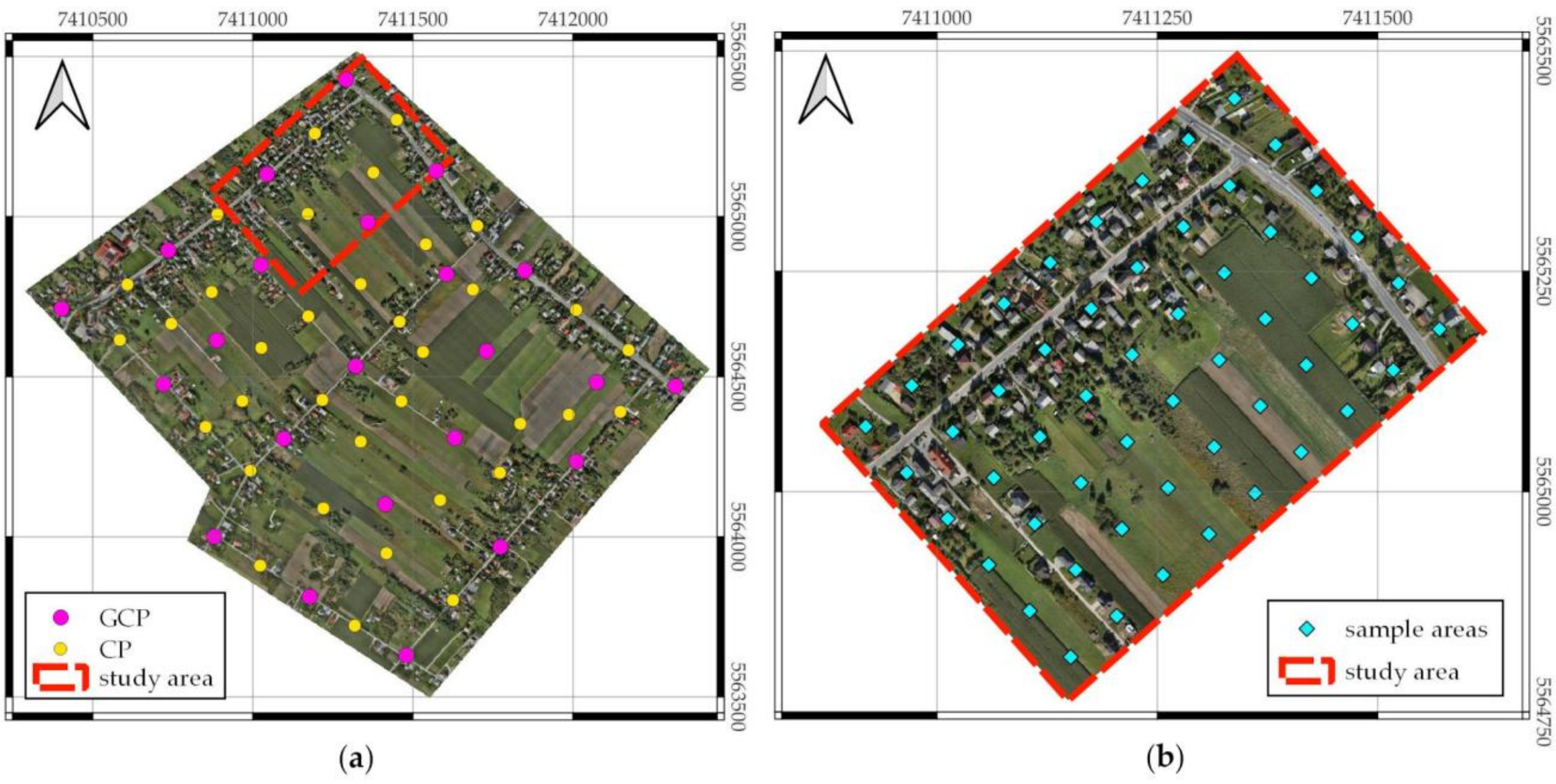

**Figure 1.** Jerzmanowice dataset: (**a**) the area covered by UAV photogrammetry missions with the locations of ground control points (GCPs) and check points (CPs) and the boundary of the study area; (**b**) the study area with the sample area locations.

Periodic UAV photogrammetry surveys were conducted in the study area. In total, 11 consecutive series were analyzed. The surveys were carried out in one-month intervals from September 2021 to August 2022. The January 2022 survey was excluded from the study due to snow coverage.

The surveys were conducted using the DJI Matrice M300 (DJI, Shenzhen, China) and the DJI Zenmuse P1 camera with the native DJI DL 35 mm F2.8 LS ASPH lens. The camera was equipped with a full-frame CMOS sensor with dimensions of 35.9 mm × 24.0 mm and a resolution of 8192 × 5460 px (45 Mpx). The DJI Matrice M300 multicopter is equipped with a global navigation satellite systems (GNSS) real-time kinematic (RTK) module, which enables the device's trajectory to be accurately determined. In all series, flight line spacings and UAV speeds were set to ensure 80% forward and 60% side image overlap, and the flight altitude was selected to ensure a ground sampling distance (GSD) of 10 mm.

During each series of photogrammetric surveys, GCPs and CPs were used (Figure 1a). Their positions were determined using the GNSS RTK technique in reference to the base stations, which were located roughly in the centers of the areas. The coordinates of the base stations were determined using the GNSS static technique in reference to the national network of reference stations (ASG-EUPOS). Agisoft Metashape Professional v.

1.6.2 build 10247 software was used to develop the photogrammetric products with a horizontal accuracy of ±1–2 cm and vertical accuracy of ±2 cm.

From the entire survey area (Figure 1a), a fragment located in the north-western part was selected for further analysis (Figure 1b). The orthomosaics were exported for this area with a spatial resolution of 10 mm/px. Then, 54 square sample areas with an area of 4 m² (40,401 pixels each) were systematically selected in a uniform grid (random sample). A reference classification was performed manually for each pixel in these samples and for each analyzed survey series, which was the basis for optimizing the classification methods' parameters (and hyperparameters). The total number of pixels for which a reference classification was performed was approximately 24 million, of which approximately 12.9% were classified as urbanized areas. The dataset obtained this way is marked as 'JERZ' later in the article.

2.1.2. Wieliczka Dataset

Data for testing classification methods were acquired using a UAV south of the city of Wieliczka (49°59′N, 20°03′E) in the Lesser Poland Voivodeship, Poland. The total area covered by UAV photogrammetry missions was 1.0 km² (Figure 2a). In total, four consecutive series were analyzed. The UAV photogrammetry surveys were carried out in quarterly intervals (the measurement dates are as follows: November 2021, February 2022, May 2022, August 2022). There are single-family houses and farm buildings concentrated along the streets. The remainder of the area is covered mainly by forests and meadows.

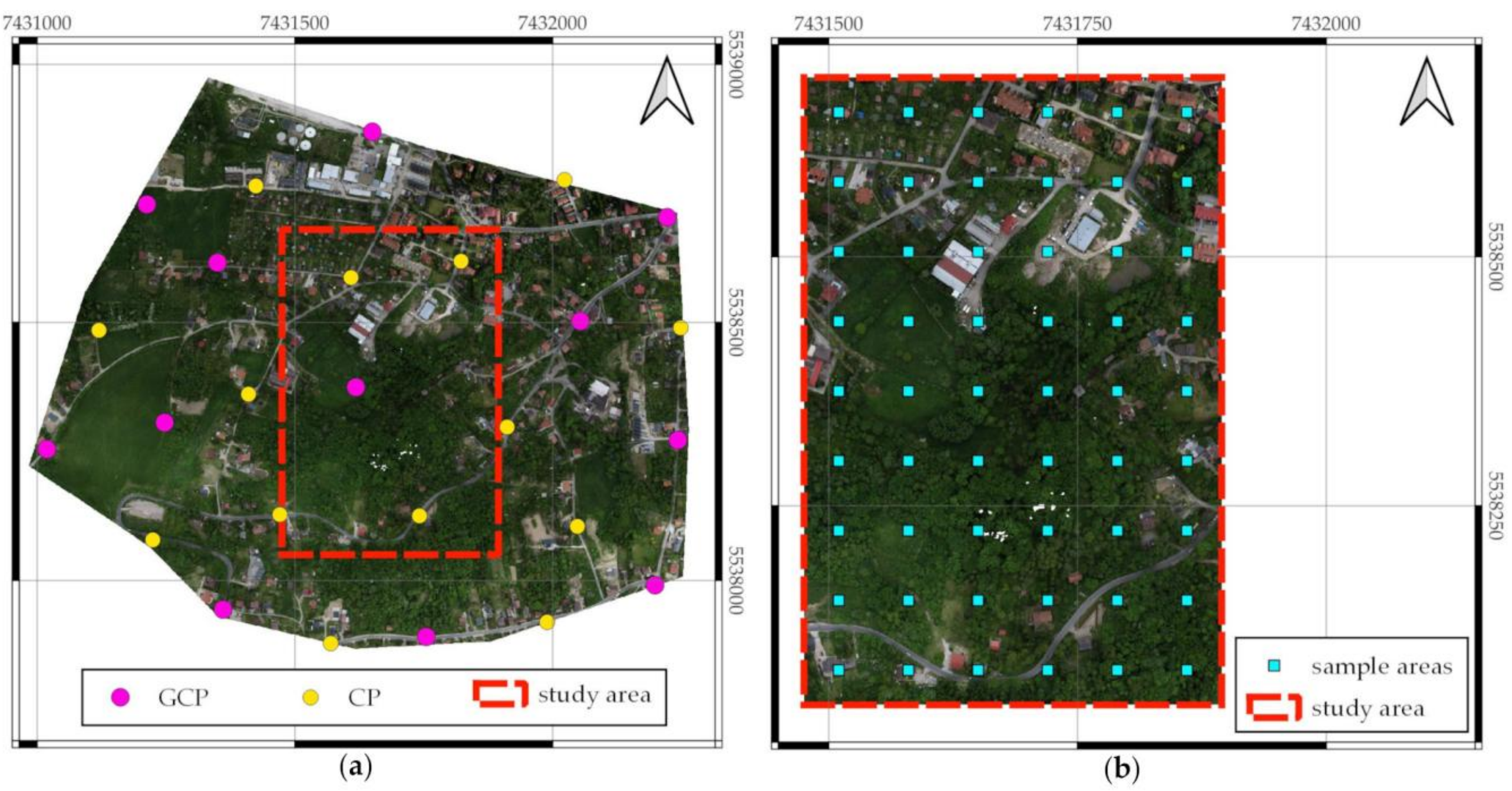

**Figure 2.** Wieliczka dataset: (**a**) the area covered by UAV photogrammetry missions with the locations of GCPs and CPs and the boundary of the study area; (**b**) the study area with the sample area locations.

The surveys were conducted using DJI Phantom 4 RTK (DJI, Shenzhen, China) and a generic camera. The camera has a 1″ RGB CMOS sensor with a resolution of 5472 × 3648 px (20 Mpx) and a lens with an 8.8 mm focal length (24 mm equivalent). The DJI Phantom 4 RTK multicopter is equipped with a GNSS RTK module. In all series, flight line spacings and UAV speeds were set to ensure 70% forward and 70% side image overlap, and a flight altitude was selected to ensure the GSD of 15 mm.

During each series of photogrammetric surveys, GCPs and CPs were used (Figure 2a). Agisoft Metashape Professional v. 1.6.2 build 10247 software was used to develop the photogrammetric products with a horizontal accuracy of ±1.5–3 cm and vertical accuracy of ±4–5 cm.

From the entire survey area (Figure 2a), a fragment located in its central part was selected for further analysis (Figure 2b). The orthomosaics were exported for this fragment with a spatial resolution of 15 mm/px. Then, 54 square sample areas with an area of 4 $m^2$ (17,956 pixels each) were systematically selected in a uniform grid (random sample). For each pixel in these samples, a reference classification was performed manually for each analyzed measurement series. The total number of pixels for which a reference classification was performed was approximately 3.8 million, of which approximately 13.3% were classified as urbanized areas. Based on these data, the accuracy of classification using individual methods was estimated. The dataset obtained this way is marked as 'WIEL' in the rest of the article.

### *2.2. Classification Methods*

This study used MATLAB (version R2022b) with the Statistics and Machine Learning Toolbox™, Deep Learning Toolbox™, and Image Processing Toolbox™ from MathWorks. Therefore, all descriptions of the tested tools are presented in accordance with the conventions adopted in these packages.

#### 2.2.1. Vegetation Indices Thresholding

This study tested 16 VIs, which are determined based on observations in the visible light range; 13 use the RGB color space, and three use the CIELab color space (Table 1). Since the classification involved division into two classes, the results of VIs processing were thresholded. The optimal threshold value was established based on the JERZ dataset, and the WIEL dataset was used only to assess the classification accuracy. This division was adopted to ensure the independence of the test dataset, as it was collected with a different sensor, under different weather conditions, and in a different area than the calibration dataset. This approach enabled a reliable classification accuracy assessment for new datasets using threshold values derived from the calibration dataset. The workflow for VI threshold calibration and assessment is presented in Figure 3. When optimizing the threshold value, the Matthews correlation coefficient (MCC) was used [37].

**Table 1.** The color VIs considered in this study.

| No. | Vegetation Index | Abbr. | Color Space | Equation | Reference |
|---|---|---|---|---|---|
| 1. | - | GBdiff | RGB | $(G - B)/(R + G + B)$ | [38] |
| 2. | Excess Green | ExG | RGB | $2g - r - b$ | [39] |
| 3. | Excess Red | ExR | RGB | $1.4r - g$ | [40] |
| 4. | Excess Blue | ExB | RGB | $1.4b - g$ | [41] |
| 5. | Excess Green Minus Excess Red | ExGR | RGB | $ExG - ExR$ | [42] |
| 6. | Excess Green Minus Excess Blue | ExGB | RGB | $ExG - ExB$ | [15] |
| 7. | Modified Excess Green Index | MExG | RGB | $1.262G - 0.884R - 0.311B$ | [43] |
| 8. | Normalized Green Red Difference Index | NGRDI | RGB | $(G - R)/(G + R)$ | [44] |
| 9. | Green Leaf Index | GLI | RGB | $(2 \times G - R - B)/(2 \times G + R + B)$ | [32] |
| 10. | Red Green Blue Vegetation Index | RGBVI | RGB | $(G^2 - B \times R)/(G^2 + B \times R)$ | [33] |
| 11. | Modified Green Red Vegetation Index | MGRVI | RGB | $(G^2 - R^2)/(G^2 + R^2)$ | [33] |
| 12. | Color Index of Vegetation Extraction | CIVE | RGB | $0.441r - 0.881g + 0.385b + 18.78745$ | [45,46] |
| 13. | Triangular Greenness Index | TGI | RGB | $G - 0.39R - 0.61B$ | [31] |
| 14. | AI | AI | CIELab | $b^* - a^*$ | [47] |
| 15. | AB | AB | CIELab | $a^*/L^*$ | [47] |
| 16. | AL | AL | CIELab | $a^*/b^*$ | [47] |

The following symbols are applied in the table: R, G, and B are standardized values (ranging from 0 to 1) for the red, green, and blue channels; r, g, and b are calculated for each pixel according to the following equations: $r = R/(R + G + B)$, $g = G/(R + G + B)$, $b = B/(R + G + B)$; $a^*$, $b^*$, and $L^*$ are the values of channels in the CIELab space.

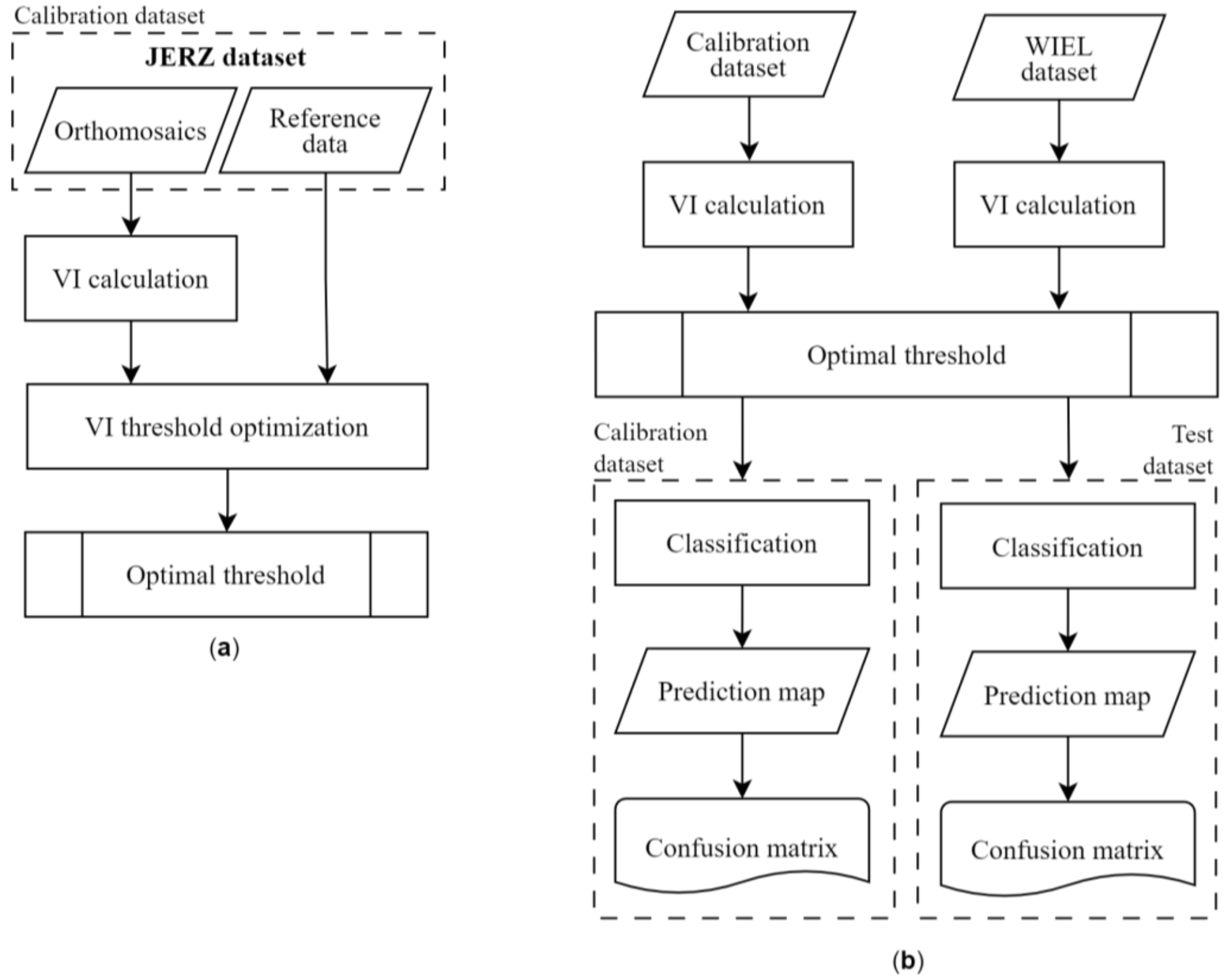

**Figure 3.** Workflow for VI threshold (**a**) calibration, and (**b**) validation and testing. The optimal threshold value was determined using the JERZ dataset (**a**), which employed orthomosaics and manually performed reference classification. Subsequently (**b**), the assessment of classification accuracy was tested for both the calibration (JERZ) and the WIEL datasets, utilizing the optimal threshold value determined in step (**a**). This ensured that the accuracies determined for the WIEL dataset were independent and unbiased.

### 2.2.2. Neural Networks

NNs were used as an alternative group of classification algorithms. Linear networks, MLP [48], and CNNs [49] were tested. All networks were trained using cross-entropy loss as an objective function:

$$L = -\sum_{i}^{C} y_i log(\hat{y}_i) \quad (1)$$

where $C$ is the number of classes ($C = 2$); $y_i$ is a binary indicator (0 or 1) if class label $i$ is the correct classification for the given observation; $\hat{y}_i$ is the predicted probability of the observation being in class $i$.

The training process ended if the validation cross-entropy loss was greater than or equal to the minimum validation cross-entropy loss computed so far, six times in a row.

The workflow for NN training, validation, and testing is presented in Figure 4. In all cases, the following pertained:

- regularization was not applied;
- three NNs of the same structure were trained, and the one with the highest MCC for the validation dataset was applied in the further tests;
- a fully connected layer with two linear units followed by softmax function was used at network outputs;
- the class for which membership probability exceeded 50% was assigned;
- the classification was conducted only for the central point of a patch;
- the patch sizes were in sets 1, 3, 5, 7, and 9, but for CNNs, only patch sizes of 5, 7, and 9 were tested;
- the training dataset was not augmented.

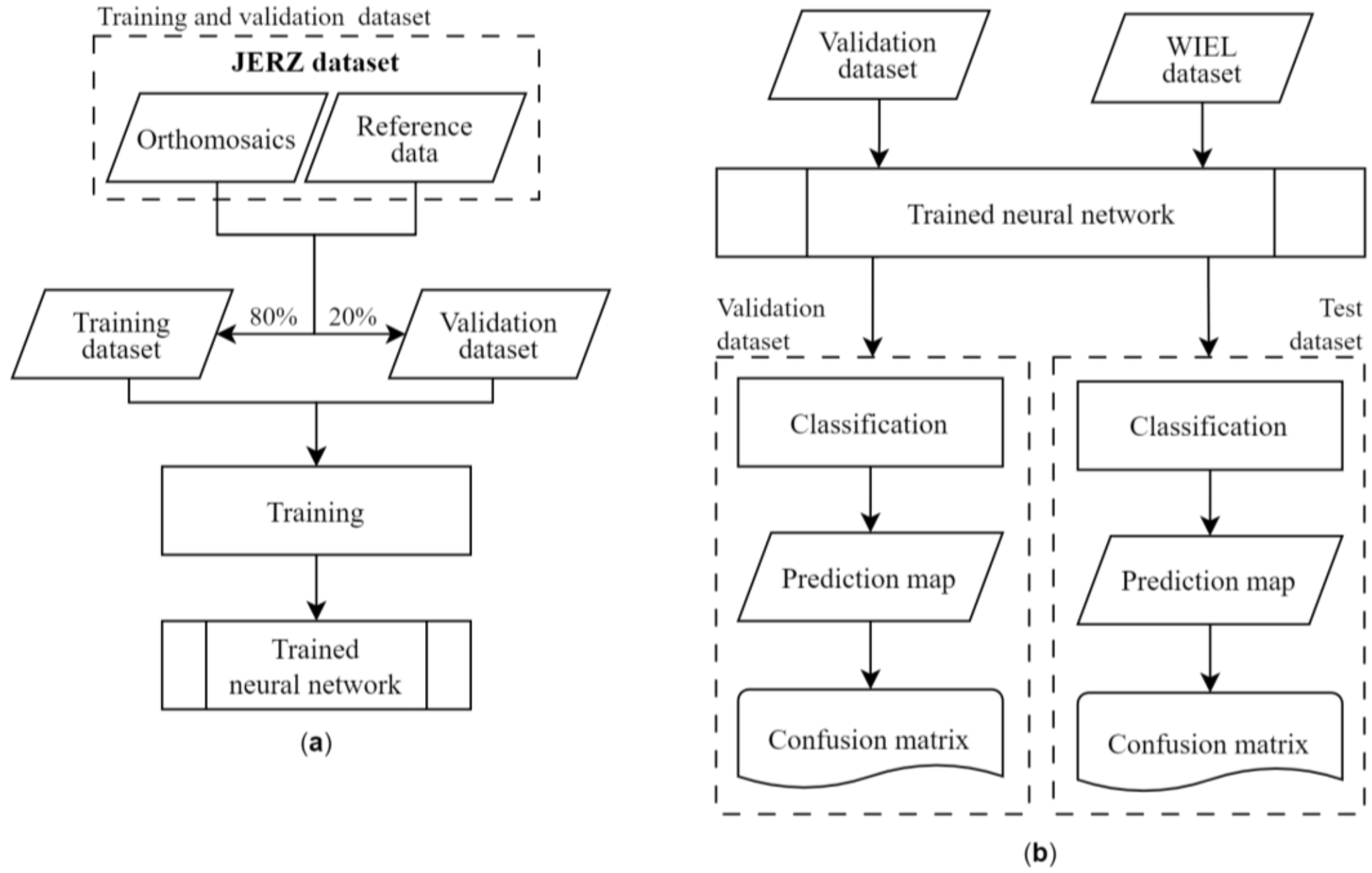

**Figure 4.** Workflow for NN (**a**) training, and (**b**) validation and testing. The training was conducted using the JERZ dataset (**a**), which employed orthomosaics and manually performed reference classification. Subsequently (**b**), classification accuracy assessment was tested for both the validation part of the JERZ dataset and the WIEL dataset. This ensured that the accuracies determined for the WIEL dataset were independent and unbiased.

The available datasets were divided as follows: 80% of the JERZ dataset was used for training, 20% for validation, and 100% of the WIEL dataset was used for testing. Similar to the VIs case, this division was intended to ensure the independence of the test dataset and provide a realistic assessment of classification accuracy.

For linear network optimization, the limited-memory Broyden–Fletcher–Goldfarb–Shanno quasi-Newton algorithm was applied [50,51]. MLP networks were trained using the scaled conjugate gradient algorithm, and CNNs were trained using the adaptive moment estimation algorithm. All MLP networks presented in this article had one hidden layer with five units, and a hyperbolic tangent sigmoid transfer function (tanh) was used. The preliminary test showed that further increasing the number of hidden layers and hidden units did not increase the performance in MLP networks. All CNNs had one convolutional layer with the five filters of size 3 with a rectified linear unit activation function. It was followed by a max pooling layer with a pool size 3 and stride 2.

### *2.3. Classification Quality Measures*

The outcome of the classification task can be described using a confusion matrix:

$$\mathbf{M} = \begin{pmatrix} \mathrm{TP} & \mathrm{FN} \\ \mathrm{FP} & \mathrm{TN} \end{pmatrix}, \tag{2}$$

where TP represents true positive (actual positives that are correctly predicted positives); FN represents false negative (actual positives that are wrongly predicted negatives); TN represents true negative (actual negatives that are correctly predicted negatives); FP represents false positive (actual negatives that are wrongly predicted positives).

Several statistical rates can be utilized to evaluate the classifications based on a confusion matrix. The most popular measure used for this task is *accuracy*:

$$accuracy = \frac{(\mathrm{TP} + \mathrm{TN})}{(\mathrm{TP} + \mathrm{TN} + \mathrm{FP} + \mathrm{FN})}. \tag{3}$$

However, when a dataset is imbalanced (the number of samples in one class is much larger than in other classes), *accuracy* cannot be considered a reliable measure because it shows overly optimistic results. In those cases, *recall* and *precision* can help evaluate the classification more adequately:

$$recall = \frac{\mathrm{TP}}{(\mathrm{TP} + \mathrm{FN})}, \tag{4}$$

$$precision = \frac{\mathrm{TP}}{(\mathrm{TP} + \mathrm{FP})}. \tag{5}$$

Intersection over Union (IoU) is another widely used evaluation metric for classification models. It measures the overlap between the predicted and the ground truth regions and is calculated as the ratio of the area of their intersection to the area of their union. The formula for calculating IoU for binary classification is as follows:

$$\mathrm{IoU} = \frac{\mathrm{TP}}{(\mathrm{TP} + \mathrm{FP} + \mathrm{FN})}. \tag{6}$$

The IoU value ranges from 0 to 1, where 0 indicates no overlap between the predicted and ground truth regions, and 1 indicates perfect overlap, representing an ideal prediction.

Many other statistical measures are utilized to evaluate the classification. The most popular ones, especially helpful for imbalanced datasets, are the F1 score and the MCC. Because the F1 score is independent of TN, and it is not symmetric for class swapping [52], MCC was used in this study (both for evaluation of classification and as optimizing criterium when determining threshold values for VIs):

$$\mathrm{MCC} = \frac{\mathrm{TP} \cdot \mathrm{TN} - \mathrm{FP} \cdot \mathrm{FN}}{\sqrt{(\mathrm{TP} + \mathrm{FP}) \cdot (\mathrm{TP} + \mathrm{FN}) \cdot (\mathrm{TN} + \mathrm{FP}) \cdot (\mathrm{TN} + \mathrm{FN})}}, \tag{7}$$

MCC is the only binary classification coefficient with a high score when the prediction performs well in all four confusion matrix categories (TP, FN, TN, and FP) [52]. Its value ranges between [−1,+1], with extreme values of −1 and +1 achieved in cases of perfect misclassification and perfect classification, respectively, while a value of 0 is the expected value for the coin tossing classifier.

## 3. Results

### *3.1. Vegetation Indices Classification Results*

After calculating the VI values for the JERZ dataset, the analyzed VIs were divided into clusters using the agglomeration algorithm [53]. The $\delta$ value calculated according to the equation was used as the metric:

$$\delta = 1 - |r|, \tag{8}$$

where $r$ is Pearson's correlation coefficient, while distances between clusters were computed using the unweighted pair group method with arithmetic mean (UPGMA). Finally, the VIs were divided into clusters, where the distances between them did not exceed a value of 0.1. The clustering results are presented on a dendrogram (Figure 5). Thus, six clusters were identified. For further analysis, the VI for which the classification results for optimal threshold yielded the highest MCC was selected from each cluster. These were as follows: ExG, ExB, NGRDI, MExG, AI, and AB. The Pearson correlation coefficients calculated for them are generally low. The exceptions here are the correlations between ExB and ExG as well as AI and MExG, which can be seen in the dendrogram.

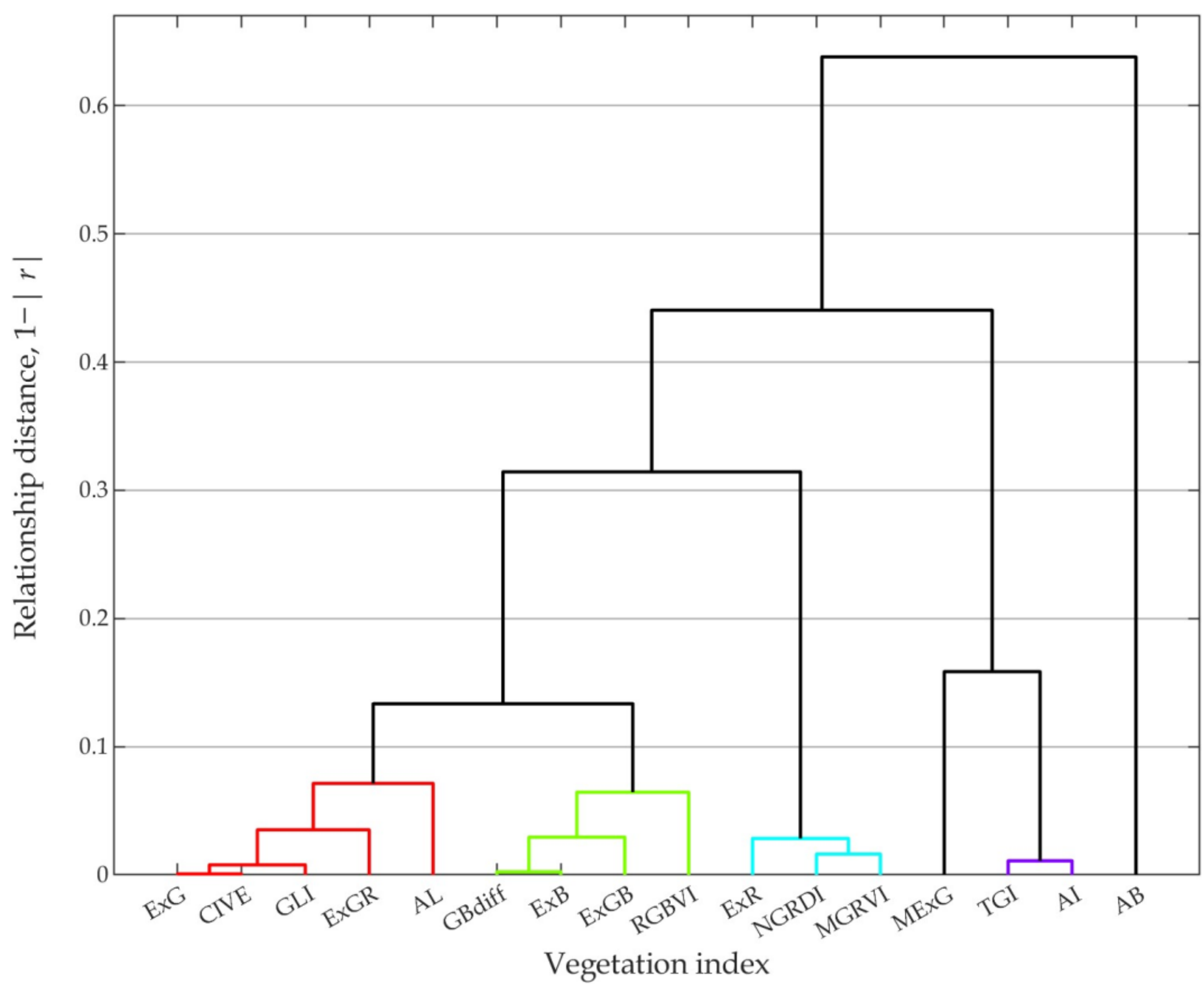

**Figure 5.** Dendrogram of associations between VIs obtained for $1 - |r|$ metric and UPGMA method using the JERZ dataset. Taking into account the 0.1 distance criterium, six clusters were identified, i.e., (1) ExG, CIVE, GLI, ExGR, AL (in red); (2) GBdiff, ExB, ExGB, RGBVI (in green); (3) ExR, NGRDI, MGRVI (in blue); (4) MExG; (5) TGI, AI (in violet); and (6) AB.

The thresholds for the analyzed VIs were chosen to maximize the MCC value for the JERZ dataset (Table 2). For the optimal thresholds, the highest MCC value was obtained for ExB, while slightly lower MCC values were obtained for AI and ExG. For the remaining VIs, the MCC values were much lower, which is also reflected in the low *precision* values. It is worth noting that the IoU values followed the same trend as the MCC. Using the same threshold values, similar results were achieved for the WIEL dataset.

**Table 2.** VI threshold values optimized for the JERZ dataset with statistical evaluation of the classification for both datasets.

| Vegetation Index | Optimal Threshold | MCC | | IoU | | *Accuracy* | | *Recall* | | *Precision* | |
|---|---|---|---|---|---|---|---|---|---|---|---|
| | | JERZ | WIEL | JERZ | WIEL | JERZ | JERZ | WIEL | JERZ | WIEL | JERZ |
| ExG | 0.020 | 0.4522 | 0.4744 | 0.3284 | 0.3512 | 0.7647 | 0.7847 | 0.8911 | 0.8812 | 0.3421 | 0.3687 |
| ExB | 0.108 | 0.6583 | 0.6099 | 0.5329 | 0.4774 | 0.9016 | 0.8700 | 0.8701 | 0.8980 | 0.5789 | 0.5048 |
| NGRDI | 0.032 | 0.1608 | 0.2863 | 0.1609 | 0.2288 | 0.3969 | 0.6436 | 0.8957 | 0.7991 | 0.1639 | 0.2427 |
| MExG | 0.078 | 0.1636 | 0.2389 | 0.1569 | 0.1875 | 0.3357 | 0.4507 | 0.9578 | 0.9580 | 0.1580 | 0.1890 |
| AI | 4.087 | 0.5113 | 0.4624 | 0.3910 | 0.3362 | 0.8346 | 0.7627 | 0.8230 | 0.9086 | 0.4269 | 0.3479 |
| AB | 0.665 | 0.2275 | 0.0910 | 0.1618 | 0.1108 | 0.8609 | 0.8063 | 0.2080 | 0.1825 | 0.4216 | 0.2201 |

Figure 6 shows histograms of VI values and the percentage of pixels classified as urbanized and non-urbanized using the optimal threshold. For ExG, NGRDI, MExG, and AI, an index value below the threshold means the pixel is classified as an urbanized area, and an index value above the threshold means the pixel is classified as a non-urbanized area. For ExB and AB, this relationship is reversed, directly resulting from how the VIs are defined. Using most VIs thresholding leads to a much larger number of pixels being classified as urbanized areas than the reference classification implies (12.9%). The variability ranges of indices are closely related to their definitions (e.g., for ExB, the variability range is from −1.0 to 1.4, while AI can reach values from −255 to +255); however, with the permissible values of indices, the optimal thresholds are generally close to zero.

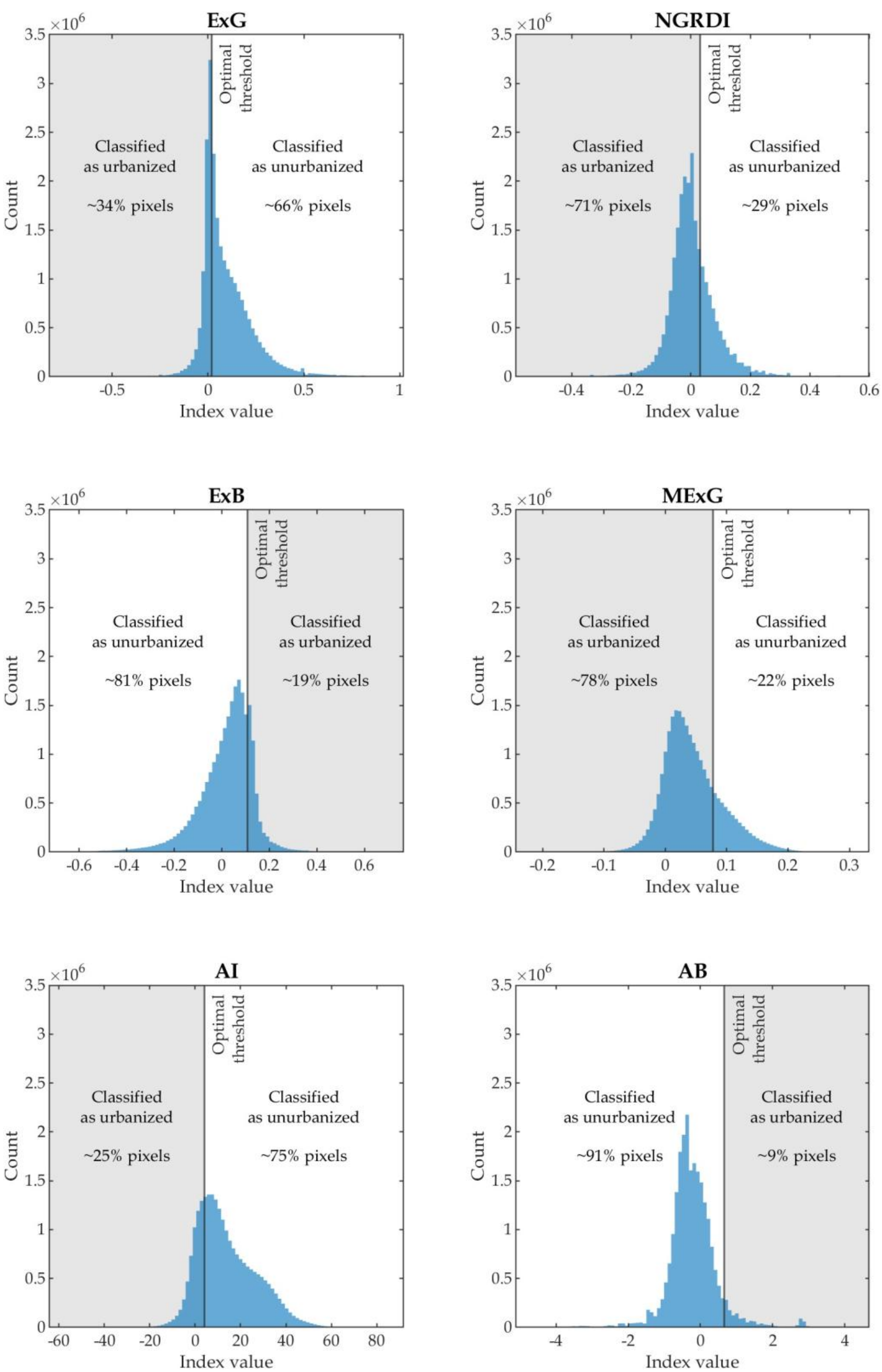

**Figure 6.** Histograms of VIs with optimal thresholds for the JERZ dataset. This figure also shows the percentage of pixels classified as urbanized and non-urbanized areas.

The influence of season on the optimal thresholds and MCC values was also analyzed. For this purpose, optimal thresholds were determined independently for each measurement series based on the JERZ dataset. For ExB, ExG, and AI, adopting the threshold determined for the entire JERZ dataset does not significantly reduce MCC values compared with values for optimal thresholds determined independently for individual measurement series (Figure 7). For ExB, which is considered the best, the optimal threshold, depending on the measurement series, ranges from 0.100 for observations from September 2021 to 0.131 for observations from December 2021. At the same time, adopting one value optimized for all series (i.e., 0.108) only slightly worsens the results compared to those obtained using threshold values optimized for specific series. In such a case, the maximum deterioration of the MCC coefficient did not exceed 0.04, with the MCC coefficient value being about 0.44 (for December 2021 series).

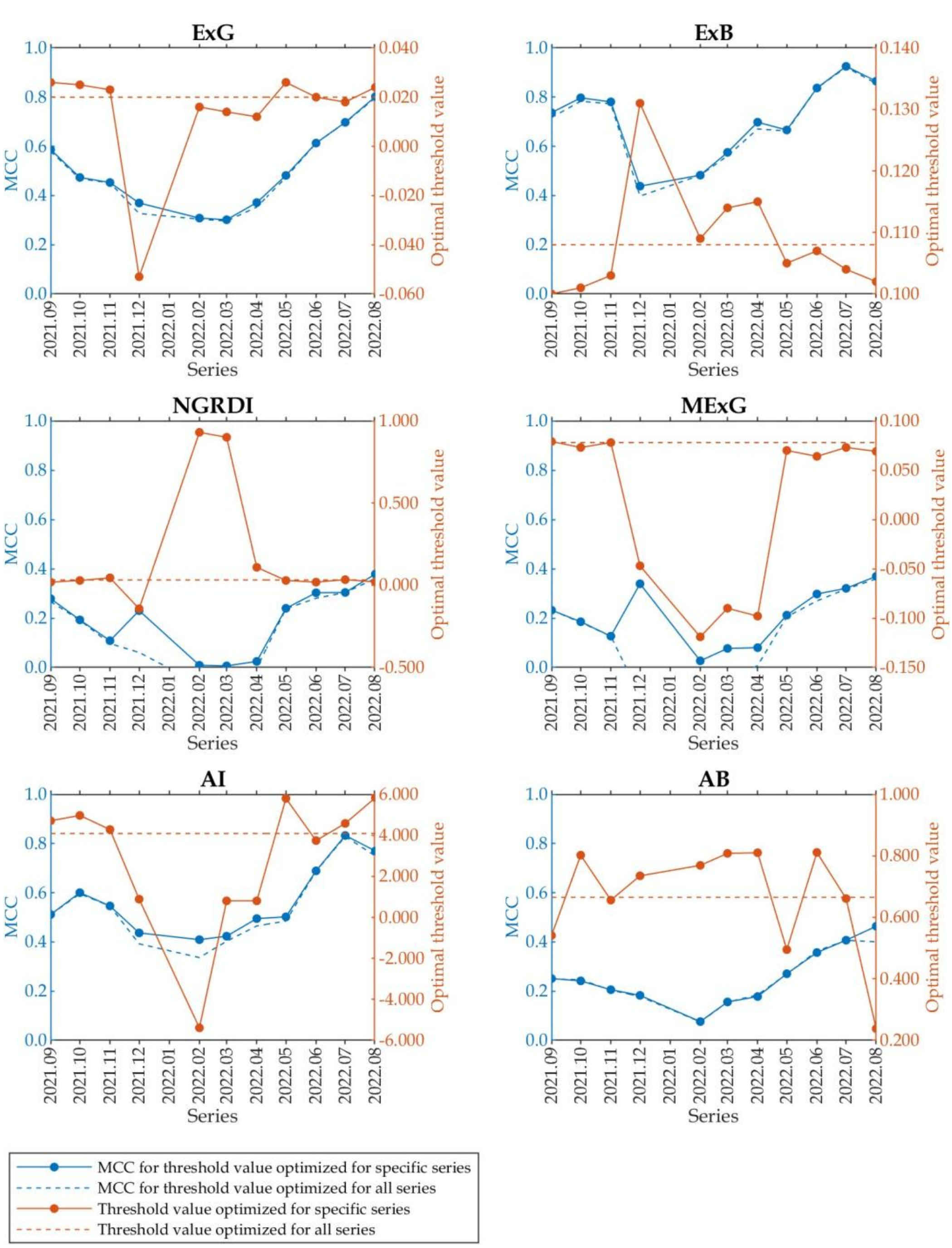

**Figure 7.** The variability in the optimal threshold depending on the survey date, with the horizontal axes displaying dates in the year.month format.

Figure 8 contains charts showing how the MCC and *accuracy* change depending on the adopted threshold value and dataset used. For ExB, the optimal threshold is the same for both analyzed datasets (JERZ and WIEL). There are some discrepancies between ExG and AI, but for ExG, the dependence of the optimal threshold on the dataset is very small, while for AI, the dataset used to optimize the threshold value has a stronger influence on the MCC value. For some VIs, *accuracy* and MCC reach their maximum for other thresholds. This is related to the dataset imbalance and the method of calculating the MCC value, which, in such a case, achieves higher values for those classification methods that classify pixels belonging to both classes with similar effectiveness. This means that methods that better classify pixels belonging to a larger class (non-urbanized areas) at the expense of errors in classifying pixels belonging to a smaller class (urbanized areas) may have relatively high *accuracy* with lower MCC. The horizontal scale of the graphs was limited to ranges for which MCC values were greater than 0.05. Taking into account this type of normalization, it is noted that VIs with relatively greater sensitivity to the adopted threshold value (ExG, ExB, AI) allowed for obtaining a higher MCC value.

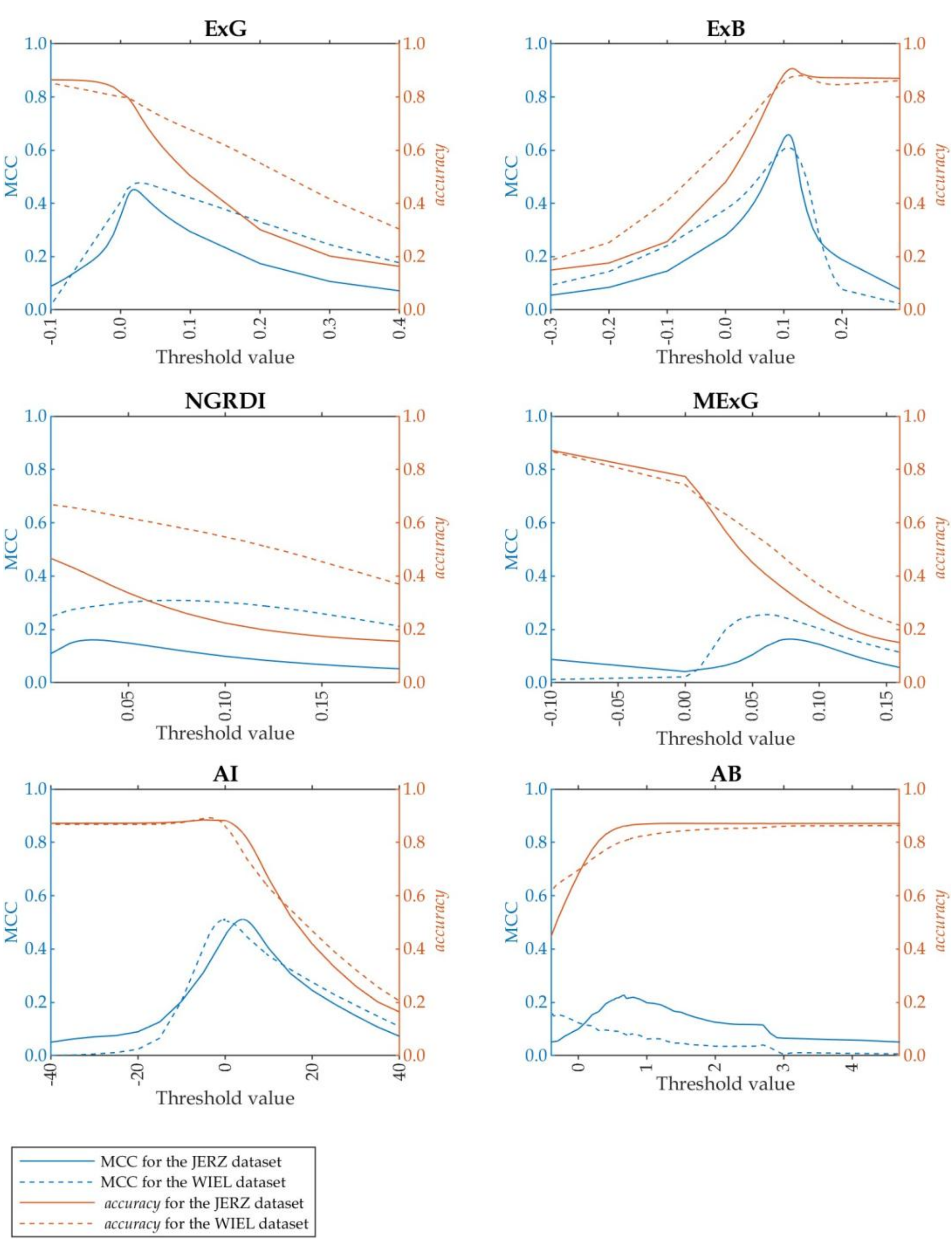

**Figure 8.** The influence of the adopted threshold value on MCC and *accuracy*.

### 3.2. Neural Networks Classification Results

Table 3 summarizes the classification results using NNs. Generally, for the validation part of the JERZ dataset (JERZ*) for all types of tested networks, the results show a noticeable improvement with increased input patch size. For this dataset (JERZ*), the best results (MCC, IoU, and *accuracy*) are obtained using CNNs, while slightly worse results are achieved for MLP and linear networks.

For the test dataset (WIEL), the MCC, IoU, and *accuracy* are similar for all network types. The largest increase in these parameters is visible when increasing the patch size from 1 px to 3 px (linear and MLP). Further increasing the patch size does not result in a large increase in accuracy. For the WIEL dataset, slightly better results are obtained using the linear network and MLP, while CNNs provide slightly poorer classification results. Generally, the highest MCC obtained for this dataset is about 0.82, IoU is about 0.73, and *accuracy* is about 0.96.

**Table 3.** Statistical evaluation of classification results for NNs.

| Network Type | Patch Size | MCC | | IoU | | *Accuracy* | | *Recall* | | *Precision* | |
|---|---|---|---|---|---|---|---|---|---|---|---|
| | | JERZ* | WIEL | JERZ* | WIEL | JERZ* | WIEL | JERZ* | WIEL | JERZ* | WIEL |
| Linear | 1 px | 0.6566 | 0.7344 | 0.5307 | 0.6162 | 0.9301 | 0.9424 | 0.6453 | 0.6967 | 0.7493 | 0.8422 |
| | 3 px | 0.7031 | 0.7926 | 0.5854 | 0.6914 | 0.9375 | 0.9539 | 0.7491 | 0.7774 | 0.7281 | 0.8621 |
| | 5 px | 0.7879 | 0.8075 | 0.6860 | 0.7098 | 0.9546 | 0.9572 | 0.8092 | 0.7868 | 0.8184 | 0.8789 |
| | 7 px | 0.7990 | 0.8163 | 0.7003 | 0.7211 | 0.9567 | 0.9591 | 0.8251 | 0.7934 | 0.8224 | 0.8878 |
| | 9 px | 0.8055 | 0.8208 | 0.7087 | 0.7267 | 0.9579 | 0.9601 | 0.8337 | 0.7953 | 0.8254 | 0.8938 |
| MLP | 1 px | 0.7040 | 0.7558 | 0.5870 | 0.6498 | 0.9370 | 0.9444 | 0.7314 | 0.7768 | 0.7484 | 0.7989 |
| | 3 px | 0.7333 | 0.7976 | 0.6172 | 0.7020 | 0.9376 | 0.9519 | 0.8540 | 0.8517 | 0.6900 | 0.7998 |
| | 5 px | 0.8324 | 0.8087 | 0.7436 | 0.7160 | 0.9623 | 0.9547 | 0.8919 | 0.8586 | 0.8173 | 0.8117 |
| | 7 px | 0.8529 | 0.8060 | 0.7716 | 0.7123 | 0.9675 | 0.9550 | 0.8947 | 0.8360 | 0.8486 | 0.8281 |
| | 9 px | 0.8457 | 0.8121 | 0.7616 | 0.7203 | 0.9655 | 0.9561 | 0.8984 | 0.8476 | 0.8334 | 0.8275 |
| CNN | 5 px | 0.8468 | 0.7927 | 0.7633 | 0.6959 | 0.9665 | 0.9503 | 0.8808 | 0.8550 | 0.8512 | 0.7889 |
| | 7 px | 0.8654 | 0.7811 | 0.7890 | 0.6817 | 0.9708 | 0.9482 | 0.8913 | 0.8332 | 0.8730 | 0.7894 |
| | 9 px | 0.8789 | 0.7808 | 0.8078 | 0.6794 | 0.9740 | 0.9503 | 0.8890 | 0.7896 | 0.8984 | 0.8296 |

Evaluation for the JERZ dataset is given only for the validation part of the dataset (JERZ*), and results for the training part of the JERZ dataset were similar (no under- or overfitting was detected). Tests were conducted for the WIEL dataset.

## 4. Discussion

Generally, the results obtained by NNs are better than those obtained by the tested VIs thresholding. The best tested VIs, i.e., ExB, achieved an MCC value for the JERZ dataset comparable with the weakest tested NNs (linear for 1 px patch size for the JERZ* dataset). Apart from this one case, and especially with an increase in patch size, NNs obtained much better classification accuracy parameters. This is not surprising, considering that the NNs were custom-developed for the classification case presented in this study, where the principal purpose of the tested VIs is quite different. However, understanding the reasons for the differences in results obtained using ExB thresholding and NNs required an in-depth review of the image classification results.

Figure 9 shows the characteristic behavior of the tested classifiers for dark parts of images. NNs tend to classify such image fragments as non-urbanized, and ExB classifies them as urbanized. As shown in Figure 9, this can lead to errors in both cases, but the much higher percentage of non-urbanized areas in the training dataset (~86.9%) and validation dataset (~87.8%) results in a statistical advantage of NNs in such cases. At the same time, in other cases, both ExB and NNs achieve similar, high classification accuracies. This applies to images showing the bare ground, vegetation, buildings, and infrastructure elements, provided that there are good lighting conditions during data acquisition, guaranteeing the high radiometric quality of the analyzed orthomosaics. The same problem was observed in another study [10], wherein the SVM model was employed to conduct and compare classification results obtained based on UAV-derived multispectral and RGB images. Regardless of the data type used, misclassification occurred in the shadowed areas, where both bare soil and dead matter areas were incorrectly identified as vegetation. Therefore, ensuring favorable atmospheric conditions with uniform scene illumination with minimal shadow impact is crucial.

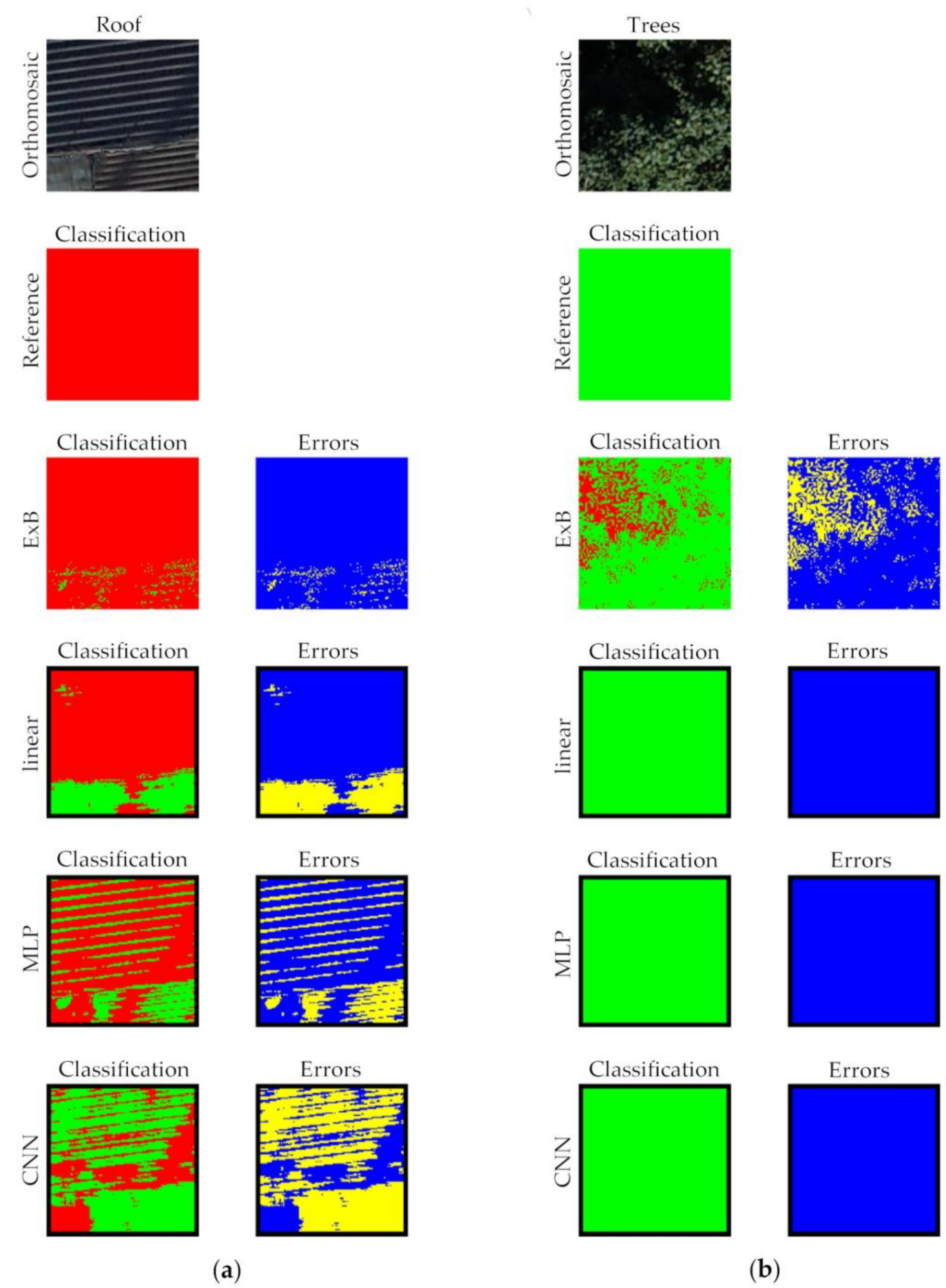

**Figure 9.** Examples of classification results for dark orthomosaic fragments: (**a**) for roof, and (**b**) for trees. The figure shows a fragment of an orthomosaic, reference classification, classification predicted by ExB and NNs (linear, MLP, and CNN), and classification errors. In the classification images, pixels classified as urbanized areas are red, and pixels classified as non-urbanized areas are green. In the error images, blue indicates correctly classified pixels, and yellow indicates misclassified pixels.

Given the limitations of the ExB index-based classifier described above, improving its accuracy prior to its implementation in an automatic displacement determination system seems likely. This can be achieved by adding a step to data processing. In this step, pixels with a brightness lower than the assumed one would be classified as a non-urbanized area regardless of the ExB value.

Simple classification into the vegetation and non-vegetation classes is often based on the VI and its threshold value that sets the breakpoint between two segmented classes. There are several automatic methods for calculating the threshold, among which Otsu's method [54] is one of the most widely used. It assumes that an image contains two classes of pixels and then calculates the optimum threshold, which maximizes the variance between them. Taking into account that the aim of the research described in the article was also to classify orthomosaics into two classes (urbanized and non-urbanized classes), the accuracy of the classification of the JERZ dataset based on the six analyzed VIs in the article and their threshold values determined using Otsu's method was verified (Table 4). Only for NGRDI, the threshold determined using Otsu's method does not differ significantly from the determined optimal threshold value (Table 2). Therefore, it has little impact on the classification accuracy, which is relatively low for NGRDI anyway. For the VI considered to be the best (ExB), the adoption of the Otsu's method-based threshold causes a significant decrease in MCC (by 0.4308), IoU (by 0.3652), and *accuracy* (by 0.5100) compared with the optimal values. It can, therefore, be concluded that Otsu's method, in this case, does not allow for determining the threshold values precisely enough for practical purposes. The optimal threshold values for VIs determined in the conducted research provide better classification results for both analyzed objects, even though these objects were observed independently using two different UAVs and cameras in different lighting conditions and resolutions.

**Table 4.** Vegetation indices threshold values determined using Otsu's method for the JERZ dataset with statistical classification evaluation.

| Vegetation Index | Otsu's Method-Based Threshold | MCC | IoU | *Accuracy* | *Recall* | *Precision* |
|---|---|---|---|---|---|---|
| ExG | 0.165 | 0.2097 | 0.1678 | 0.3615 | 0.9972 | 0.1678 |
| ExB | −0.031 | 0.2275 | 0.1677 | 0.3916 | 0.9970 | 0.1747 |
| NGRDI | 0.035 | 0.1600 | 0.1599 | 0.3854 | 0.9063 | 0.1626 |
| MExG | 0.059 | 0.1327 | 0.1551 | 0.4114 | 0.8370 | 0.1599 |
| AI | 16.887 | 0.2814 | 0.1985 | 0.4802 | 0.9973 | 0.1986 |
| AB | −0.134 | 0.0826 | 0.1402 | 0.6095 | 0.4935 | 0.1638 |

Due to the different resolutions of the orthomosaics in the JERZ and WIEL datasets for NNs, an analysis was performed for two approaches to constructing patches. In the first, for which the results are presented in Table 3, the patches provided at the network input had the same pixel size for both datasets (JERZ and WIEL), meaning their physical area was different. The second approach was based on orthomosaic subsampling from the WIEL dataset so that the physical patch area for both datasets was the same. In both cases, classification accuracy characteristics were similar, with a slight advantage for the non-subsampling approach (Table 3).

Attempts were also made to use NNs receiving as inputs: (a) values from RGB channels extended by the ExB index value (as the fourth channel), as well as (b) pixel brightness values in the first channel and ExB index values in the second channel. In both cases, the results were no better than those obtained using NNs, whose inputs included only values from RGB channels. Taking this into account, it should be recognized that, unlike many other classifiers based on ML methods known in the literature, the relatively simplest solution allowed for obtaining the best classification accuracy characteristics.

Previous research results on LULC classification based on UAV RGB images have shown that classification accuracy can reach up to 98% [27]. The methods evaluated in this study achieved a classification accuracy of 87% using the best VI and 96% using the best NN. The classification accuracies reported in the different studies should not be directly compared to determine the best model, as the datasets used in each study differed.

However, these values provide an approximate benchmark for the expected level of classification accuracy.

Unlike previous studies in this field [10,19,22,24,26,27,30,34], the test dataset was acquired in a different area than the training/calibration dataset, using a distinct UAV platform in four measurement series conducted at three-month intervals (in different seasons). This approach provided a comprehensive and reliable assessment of the methods' accuracy, ensuring satisfactory classification accuracy for varied data and study areas. This was achieved by incorporating data from 11 measurement series, conducted monthly, into the training/calibration process. Notably, this approach has not been previously utilized in developing LULC classifiers based on UAV RGB images. Consequently, the trained NNs and the determined optimal VI's thresholds remain robust across varying vegetation states resulting from seasonal variability.

Our research concerned extremely simple solutions compared to recent developments in LULC classifiers. Current trends emphasize the successful use of DL methods. However, a major challenge associated with such models is the requirement for large datasets with reference classification. While several open datasets are available, the images in these datasets have much lower spatial resolution than those considered in this research. Additionally, most state-of-the-art LULC classifiers are developed with satellite imagery in mind and utilize a wider spectrum of bands beyond visible light. Acquiring such data using UAV is notably more challenging and costly than obtaining RGB images alone. Consequently, UAV data often have much lower spectral resolution. Even if a UAV is equipped with a multispectral sensor, the spatial resolution of the data would be significantly lower compared to using an RGB camera.

Acquiring both multispectral and high-resolution RGB images necessitates using a UAV equipped with dual sensors or conducting two separate UAV missions with different payloads. Practically, either approach considerably increases fieldwork, potentially limiting the feasibility of systems where classification is only an optional step in UAV RGB data processing [12,35]. Therefore, if classification accuracy is secondary to the necessity for high-resolution RGB images, processing should prioritize RGB bands, possibly augmented by elevation data and local image features of a statistical nature, to improve classification accuracy [21,22,24].

As this research is related to constructing a system aimed at determining displacements based on UAV photogrammetry [12,35], the results primarily served to classify a monitored region into urbanized and non-urbanized areas (i.e., classify registration windows). Additionally, horizontal displacements are determined within the system using weighted normalized cross-correlation, where high-resolution classification is utilized as weights. An alternative approach would require reformulation of the problem, i.e., reducing the spatial resolution of classification for the weights matrix and classification of the registration windows (urbanized vs. non-urbanized) while retaining the full RGB image spatial resolution to the core of the image registration and displacement determination. This would allow easier (and less time-consuming) acquisition of reference classification and consequently enable the application of more complex DL methods. This could potentially also reduce the number of outliers in determined displacements without significantly increasing processing time. However, further experiments are necessary to validate such alternative approaches.

## 5. Conclusions

The research developed classification methods accurate enough to support a system determining ground surface displacements using high-resolution photogrammetric products. Better classification results were obtained using NNs trained specifically for this task. The accuracy of these models increases with the image patch size fed to their inputs, and for test data, the highest MCC values were obtained using a linear network and slightly lower values were obtained for MLP and CNN networks.

Among the tested VIs, the ExB achieved the best classification into urbanized and non-urbanized areas. Although classification accuracy obtained using it is lower than that obtained using NNs, its use is sufficient for the given task. This, combined with its simplicity of construction and the associated processing speed, makes it a valuable classifier.

**Author Contributions:** Conceptualization, E.P. (Edyta Puniach), W.G., P.Ć., and E.P. (Elżbieta Pastucha); methodology, E.P. (Edyta Puniach) and W.G.; software, E.P. (Edyta Puniach) and W.G.; validation, E.P. (Edyta Puniach); formal analysis, E.P. (Edyta Puniach) and W.G.; investigation, E.P. (Edyta Puniach) and W.G.; resources, P.Ć. and K.S.; data curation, E.P. (Edyta Puniach), P.Ć., and K.S.; writing—original draft preparation, E.P. (Edyta Puniach), W.G., and E.P. (Elżbieta Pastucha); writing—review and editing, E.P. (Edyta Puniach), W.G., and E.P. (Elżbieta Pastucha); visualization, E.P. (Edyta Puniach) and W.G.; supervision, W.G.; project administration, P.Ć.; funding acquisition, E.P. (Edyta Puniach), W.G., and P.Ć. All authors have read and agreed to the published version of the manuscript.

**Funding:** This research project was supported by the program "Excellence Initiative—Research University" for the AGH University of Krakow and statutory research of AGH University of Krakow, Faculty of Geo-Data Science, Geodesy, and Environmental Engineering [grant number 16.16.150.545].

**Data Availability Statement:** The data that support this study's findings are available in the RODBUK repository at https://doi.org/10.58032/AGH/9TJA3T.

**Conflicts of Interest:** The authors declare no conflicts of interest.

## References

1. Arpitha, M.; Ahmed, S.A.; Harishnaika, N. Land use and land cover classification using machine learning algorithms in google earth engine. *Earth Sci. Inform.* **2023**, *16*, 3057–3073. https://doi.org/10.1007/s12145-023-01073-w.
2. Moharram, M.A.; Sundaram, D.M. Land use and land cover classification with hyperspectral data: A comprehensive review of methods, challenges and future directions. *Neurocomputing* **2023**, *536*, 90–113. https://doi.org/10.1016/j.neucom.2023.03.025.
3. Wang, Y.; Sun, Y.; Cao, X.; Wang, Y.; Zhang, W.; Cheng, X. A review of regional and Global scale Land Use/Land Cover (LULC) mapping products generated from satellite remote sensing. *ISPRS J. Photogramm. Remote Sens.* **2023**, *206*, 311–334. https://doi.org/10.1016/j.isprsjprs.2023.11.014.
4. Tokarczyk, P.; Leitao, J.P.; Rieckermann, J.; Schindler, K.; Blumensaat, F. High-quality observation of surface imperviousness for urban runoff modelling using UAV imagery. *Hydrol. Earth Syst. Sci.* **2015**, *19*, 4215–4228. https://doi.org/10.5194/hess-19-4215-2015.
5. Liao, W.; Deng, Y.; Li, M.; Sun, M.; Yang, J.; Xu, J. Extraction and Analysis of Finer Impervious Surface Classes in Urban Area. *Remote Sens.* **2021**, *13*, 459. https://doi.org/10.3390/rs13030459.
6. Shao, Z.; Cheng, T.; Fu, H.; Li, D.; Huang, X. Emerging Issues in Mapping Urban Impervious Surfaces Using High-Resolution Remote Sensing Images. *Remote Sens.* **2023**, *15*, 2562. https://doi.org/10.3390/rs15102562.
7. Alvarez-Vanhard, E.; Houet, T.; Mony, C.; Lecoq, L.; Corpetti, T. Can UAVs fill the gap between in situ surveys and satellites for habitat mapping? *Remote Sens. Environ.* **2020**, *243*, 111780. https://doi.org/10.1016/j.rse.2020.111780.
8. Gokool, S.; Mahomed, M.; Brewer, K.; Naiken, V.; Clulow, A.; Sibanda, M.; Mabhaudhi, T. Crop mapping in smallholder farms using unmanned aerial vehicle imagery and geospatial cloud computing infrastructure. *Heliyon* **2024**, *10*, e26913. https://doi.org/10.1016/j.heliyon.2024.e26913.
9. Mollick, T.; Azam, M.G.; Karim, S. Geospatial-based machine learning techniques for land use and land cover mapping using a high-resolution unmanned aerial vehicle image. *Remote Sens. Appl. Soc. Environ.* **2023**, *29*, 100859. https://doi.org/10.1016/j.rsase.2022.100859.
10. Furukawa, F.; Laneng, L.A.; Ando, H.; Yoshimura, N.; Kaneko, M.; Morimoto, J. Comparison of RGB and multispectral unmanned aerial vehicle for monitoring vegetation coverage changes on a landslide area. *Drones* **2021**, *5*, 97. https://doi.org/10.3390/drones5030097.
11. Ćwiąkała, P.; Gruszczyński, W.; Stoch, T.; Puniach, E.; Mrocheń, D.; Matwij, W.; Matwij, K.; Nędzka, M.; Sopata, P.; Wójcik, A. UAV applications for determination of land deformations caused by underground mining. *Remote Sens.* **2020**, *12*, 1733. https://doi.org/10.3390/rs12111733.
12. Puniach, E.; Gruszczyński, W.; Ćwiąkała, P.; Matwij, W. Application of UAV-based orthomosaics for determination of horizontal displacement caused by underground mining. *ISPRS J. Photogramm. Remote Sens.* **2021**, *174*, 282–303. https://doi.org/10.1016/j.isprsjprs.2021.02.006.
13. da Silva, V.S.; Salami, G.; da Silva, M.I.O.; Silva, E.A.; Monteiro Junior, J.J.; Alba, E. Methodological evaluation of vegetation indexes in land use and land cover (LULC) classification. *Geol. Ecol. Landsc.* **2020**, *4*, 159–169. https://doi.org/10.1080/24749508.2019.1608409.
14. Rouse, J.W.; Haas, R.H.; Schell, J.A.; Deering, D.W. Monitoring vegetation systems in the Great Plains with ERTS. In Proceedings of Third Symposium of ERTS, Greenbelt, MD, USA, 1974; pp. 309–317.
15. Cho, W.; Iida, M.; Suguri, M.; Masuda, R.; Kurita, H. Vision-based uncut crop edge detection for automated guidance of head-feeding combine. *Eng. Agric. Environ. Food* **2014**, *7*, 97–102. https://doi.org/10.1016/j.eaef.2013.12.010.

16. Maxwell, A.E.; Warner, T.A.; Fang, F. Implementation of machine-learning classification in remote sensing: An applied review. *Int. J. Remote Sens.* **2018**, *39*, 2784–2817. 10.1080/01431161.2018.1433343.
17. Talukdar, S.; Singha, P.; Mahato, S.; Pal, S.; Liou, Y.-A.; Rahman, A. Land-Use Land-Cover Classification by Machine Learning Classifiers for Satellite Observations—A Review. *Remote Sens.* **2020**, *12*, 1135. https://doi.org/10.3390/rs12071135.
18. Memduhoğlu, A. Identifying impervious surfaces for rainwater harvesting feasibility using unmanned aerial vehicle imagery and machine learning classification. *Adv. GIS* **2023**, *3*, 1–6.
19. Lu, T.; Wan, L.; Qi, S.; Gao, M. Land Cover Classification of UAV Remote Sensing Based on Transformer–CNN Hybrid Architecture. *Sensors* **2023**, *23*, 5288. https://doi.org/10.3390/s23115288.
20. Zhao, S.; Tu, K.; Ye, S.; Tang, H.; Hu, Y.; Xie, C. Land Use and Land Cover Classification Meets Deep Learning: A Review. *Sensors* **2023**, *23*, 8966. https://doi.org/10.3390/s23218966.
21. Aryal, J.; Sitaula, C.; Frery, A.C. Land use and land cover (LULC) performance modeling using machine learning algorithms: A case study of the city of Melbourne, Australia. *Sci. Rep.* **2023**, *13*, 13510. https://doi.org/10.1038/s41598-023-40564-0.
22. Chen, J.; Chen, Z.; Huang, R.; You, H.; Han, X.; Yue, T.; Zhou, G. The Effects of Spatial Resolution and Resampling on the Classification Accuracy of Wetland Vegetation Species and Ground Objects: A Study Based on High Spatial Resolution UAV Images. *Drones* **2023**, *7*, 61. https://doi.org/10.3390/drones7010061.
23. Gibril, M.B.A.; Kalantar, B.; Al-Ruzouq, R.; Ueda, N.; Saeidi, V.; Shanableh, A.; Mansor, S.; Shafri, H.Z.M. Mapping Heterogeneous Urban Landscapes from the Fusion of Digital Surface Model and Unmanned Aerial Vehicle-Based Images Using Adaptive Multiscale Image Segmentation and Classification. *Remote Sens.* **2020**, *12*, 1081. https://doi.org/10.3390/rs12071081.
24. Li, Y.; Deng, T.; Fu, B.; Lao, Z.; Yang, W.; He, H.; Fan, D.; He, W.; Yao, Y. Evaluation of Decision Fusions for Classifying Karst Wetland Vegetation Using One-Class and Multi-Class CNN Models with High-Resolution UAV Images. *Remote Sens.* **2022**, *14*, 5869. https://doi.org/10.3390/rs14225869.
25. Park, G.; Park, K.; Song, B.; Lee, H. Analyzing Impact of Types of UAV-Derived Images on the Object-Based Classification of Land Cover in an Urban Area. *Drones* **2022**, *6*, 71. https://doi.org/10.3390/drones6030071.
26. Öztürk, M.Y.; Çölkesen, I. The impacts of vegetation indices from UAV-based RGB imagery on land cover classification using ensemble learning. *Mersin Photogramm. J.* **2021**, *3*, 41–47. https://doi.org/10.53093/mephoj.943347.
27. Al-Najjar, H.A.; Kalantar, B.; Pradhan, B.; Saeidi, V.; Halin, A.A.; Ueda, N.; Mansor, S. Land cover classification from fused DSM and UAV images using convolutional neural networks. *Remote Sens.* **2019**, *11*, 1461. https://doi.org/10.3390/rs11121461.
28. Elamin, A.; El-Rabbany, A. UAV-Based Multi-Sensor Data Fusion for Urban Land Cover Mapping Using a Deep Convolutional Neural Network. *Remote Sens.* **2022**, *14*, 4298. https://doi.org/10.3390/rs14174298.
29. Abdolkhani, A.; Attarchi, S.; Alavipanah, S.K. A new classification scheme for urban impervious surface extraction from UAV data. *Earth Sci. Inform.* **2024**. https://doi.org/10.1007/s12145-024-01430-3.
30. Fan, C.L. Ground surface structure classification using UAV remote sensing images and machine learning algorithms. *Appl. Geomat.* **2023**, *15*, 919–931. https://doi.org/10.1007/s12518-023-00530-x.
31. Concepcion, R.S.; Lauguico, S.C.; Tobias, R.R.; Dadios, E.P.; Bandala, A.A.; Sybingco, E. Estimation of photosynthetic growth signature at the canopy scale using new genetic algorithm-modified visible band triangular greenness index. In Proceedings of 2020 International Conference on Advanced Robotics and Intelligent Systems (ARIS), Taipei, Taiwan, 19–21 August 2020; pp. 1–6. https://doi.org/10.1109/ARIS50834.2020.9205787.
32. Louhaichi, M.; Borman, M.M.; Johnson, D.E. Spatially located platform and aerial photography for documentation of grazing impacts on wheat. *Geocarto Int.* **2001**, *16*, 65–70. https://doi.org/10.1080/10106040108542184.
33. Bendig, J.; Yu, K.; Aasen, H.; Bolten, A.; Bennertz, S.; Broscheit, J.; Gnyp, M.L.; Bareth, G. Combining UAV-based plant height from crop surface models, visible, and near infrared vegetation indices for biomass monitoring in barley. *Int. J. Appl. Earth Obs. Geoinf.* **2015**, *39*, 79–87. https://doi.org/10.1016/j.jag.2015.02.012.
34. Lu, N.; Zhou, J.; Han, Z.; Li, D.; Cao, Q.; Yao, X.; Tian, Y.; Zhu, Y.; Cao, W.; Cheng, T. Improved estimation of aboveground biomass in wheat from RGB imagery and point cloud data acquired with a low-cost unmanned aerial vehicle system. *Plant Methods* **2019**, *15*, 17. https://doi.org/10.1186/s13007-019-0402-3.
35. Puniach, E.; Gruszczyński, W.; Stoch, T.; Mrocheń, D.; Ćwiąkała, P.; Sopata, P.; Pastucha, E.; Matwij, W. Determination of the coefficient of proportionality between horizontal displacement and tilt change using UAV photogrammetry. *Eng. Geol.* **2023**, *312*, 106939. https://doi.org/10.1016/j.enggeo.2022.106939.
36. Gruszczyński, W.; Puniach, E.; Ćwiąkała, P.; Matwij, W. Correction of low vegetation impact on UAV-derived point cloud heights with U-Net networks. *IEEE Trans. Geosci. Remote Sens.* **2022**, *60*, 3057272. https://doi.org/10.1109/TGRS.2021.3057272.
37. Matthews, B.W. Comparison of the predicted and observed secondary structure of T4 phage lysozyme. *Biochim. Biophys Acta (BBA) Protein Struct.* **1975**, *405*, 442–451. https://doi.org/10.1016/0005-279590109-9.
38. Kawashima, S.; Nakatani, M. An algorithm for estimating chlorophyll content in leaves using a video camera. *Ann. Bot.* **1998**, *81*, 49–54. https://doi.org/10.1006/anbo.1997.0544.
39. Woebbecke, D.M.; Meyer, G.E.; Von Bargen, K.; Mortensen, D.A. Color indices for weed identification under various soil, residue, and lighting conditions. *Trans. ASAE* **1995**, *38*, 259–269. https://doi.org/10.13031/2013.27838.
40. Meyer, G.E.; Hindman, T.W.; Laksmi, K. Machine vision detection parameters for plant species identification. *Precis. Agric. Biol. Qual.* **1999**, *3543*, 327–335. https://doi.org/10.1117/12.336896.
41. Mao, W.; Wang, Y.; Wang, Y. Real-time detection of between-row weeds using machine vision. In Proceedings of 2003 ASAE Annual Meeting, Las Vegas, Nevada, USA, 27–30 July 2003; American Society of Agricultural and Biological Engineers: St. Joseph, MI, USA, 2003. https://doi.org/10.13031/2013.15381.

42. Meyer, G.E.; Neto, J.C. Verification of color vegetation indices for automated crop imaging applications. *Comput. Electron. Agric.* **2008**, *63*, 282–293. https://doi.org/10.1016/j.compag.2008.03.009.
43. Burgos-Artizzu, X.P.; Ribeiro, A.; Guijarro, M.; Pajares, G. Real-time image processing for crop/weed discrimination in maize fields. *Comput. Electron. Agric.* **2011**, *75*, 337–346. https://doi.org/10.1016/j.compag.2010.12.011.
44. Tucker, C.J. Red and photographic infrared linear combinations for monitoring vegetation. *Remote Sens. Environ.* **1979**, *8*, 127–150. https://doi.org/10.1016/0034-425790013-0.
45. Tosaka, N.; Hata, S.; Okamoto, H.; Takai, M. Automatic thinning mechanism of sugar beets, 2: Recognition of sugar beets by image color information. *J. Jpn. Soc. Agric. Mach.* **1998**, *60*, 75–82. https://doi.org/10.11357/jsam1937.60.2_75.
46. Kataoka, T.; Kaneko, T.; Okamoto, H.; Hata, S. Crop growth estimation system using machine vision. In Proceedings of 2003 IEEE/ASME International Conference on Advanced Intelligent Mechatronics (AIM 2003), Kobe, Japan, 20–24 July 2003; pp. b1079–b1083. https://doi.org/10.1109/AIM.2003.1225492.
47. Wahono, W.; Indradewa, D.; Sunarminto, B.H.; Haryono, E.; Prajitno, D. CIE L* a* b* color space based vegetation indices derived from unmanned aerial vehicle captured images for chlorophyll and nitrogen content estimation of tea (Camellia sinensis L. Kuntze) leaves. *Ilmu Pertan. (Agric. Sci.)* **2019**, *4*, 46–51. https://doi.org/10.22146/ipas.40693.
48. Bishop, M.C. *Pattern Recognition and Machine Learning*; Springer: New York, NY, USA, 2006.
49. LeCun, Y.; Bengio, Y.; Hinton, G. Deep learning. *Nature* **2015**, *521*, 436–444.
50. Nocedal, J.; Updating quasi-Newton matrices with limited storage. *Math. Comput.* **1980**, *35*, 773–782.
51. Liu, D.C.; Nocedal, J. On the limited memory BFGS method for large scale optimization. *Math. Program.* **1989**, *45*, 503–528.
52. Chicco, D.; Jurman, G. The advantages of the Matthews correlation coefficient (MCC) over F1 score and accuracy in binary classification evaluation. *BMC Genom.* **2020**, *21*, 6. https://doi.org/10.1186/s12864-019-6413-7.
53. Sneath, P.H.; Sokal, R.R. Numerical taxonomy. In *The Principles and Practice of Numerical Classification*, 1st ed.; W. H. Freeman: San Francisco, CA, USA, 1973.
54. Otsu, N. A threshold selection method from gray-level histograms. *IEEE Trans. Syst. Man Cybern. Syst.* **1979**, *9*, 62–66. https://doi.org/10.1109/TSMC.1979.4310076.